\documentclass[sigconf,authorversion=true]{acmart}

\usepackage{booktabs} 
\usepackage{algorithmic}
\usepackage{algorithm2e}
\usepackage{graphicx,subfig}

\begin{document}
\title{Evolution of a Functionally Diverse Swarm via a Novel Decentralised Quality-Diversity Algorithm }


\author{Emma Hart}
\affiliation{%
  \institution{Edinburgh Napier University}
  \city{Edinburgh, Scotland, UK} 
  \postcode{EH10 5DT}
}
\email{e.hart@napier.ac.uk}

\author{Andreas S.W. Steyven}
\orcid{0000-0001-9570-3697}
\affiliation{%
 \institution{Edinburgh Napier University}
  \city{Edinburgh, Scotland, UK} 
  \postcode{EH10 5DT}
}
\email{a.steyven@napier.ac.uk}

\author{Ben Paechter}
\affiliation{%
\institution{Edinburgh Napier University}
  \city{Edinburgh, Scotland, UK} 
  \postcode{EH10 5DT}
}
\email{b.paechter@napier.ac.uk}

\copyrightyear{2018} 
\acmYear{2018} 
\setcopyright{acmcopyright}
\acmConference[GECCO '18]{Genetic and Evolutionary Computation Conference}{July 15--19, 2018}{Kyoto, Japan}
\acmBooktitle{GECCO '18: Genetic and Evolutionary Computation Conference, July 15--19, 2018, Kyoto, Japan}
\acmPrice{15.00}
\acmDOI{10.1145/3205455.3205481}
\acmISBN{978-1-4503-5618-3/18/07}

\renewcommand{\shorttitle}{Evolution of a swarm via a decentralised QD algorithm}

\begin{abstract}
The presence of functional
diversity within a group has been demonstrated to lead to greater robustness, higher performance and increased problem-solving ability in a broad range of studies that includes insect groups, human groups and swarm robotics. Evolving group diversity however has proved challenging within Evolutionary Robotics, requiring reproductive isolation and careful attention to population size and selection mechanisms. To tackle this issue, we introduce a novel, decentralised, variant of the MAP-Elites illumination algorithm which is hybridised with a well-known distributed evolutionary algorithm (mEDEA). The algorithm simultaneously evolves multiple diverse behaviours for multiple robots, with respect to a simple token-gathering task.  Each robot in the swarm maintains a local archive defined by two pre-specified functional traits which is shared with robots it come into contact with. We investigate four different strategies for sharing, exploiting and combining local archives and compare  results to mEDEA. Experimental results show  that in contrast to previous claims, it is possible to evolve a functionally diverse swarm without geographical isolation, and that the new method outperforms mEDEA in terms of the diversity, coverage and precision of the evolved swarm.
\end{abstract}

%
%



\maketitle

\section{Introduction}
\label{sec:intro}

In both natural and artificial systems, the benefits of
{\em functional diversity} within a species or group is well understood.  In ecology for example, functional diversity within bee pollinators (such as flower height preference, daily time of flower visitation and within-flower behaviour) has been shown to lead to increased pollination rates \cite{hoehn2008functional} which in turn sustain the bee population. Within groups of humans, functional diversity reflects differences in how people represent problems and how they go about solving them. A 
number of studies indicate the benefit of functional diversity within a group  with studies using agent-based modelling suggesting that  collective diversity trumps    individual ability under certain conditions in groups of problem solvers \cite{Hong16385}. In Computer Science, a body of theory supports the  requirement for behavioural diversity within ensemble classifiers \cite{breiman2001random} while more recent work suggests that diversity may also benefit ensembles of optimisation algorithms \cite{Hart17}.
A current review of diversity preservation in EAs is given in \cite{Squillero2016}.


We hypothesise that the presence of functional diversity within a robotic swarm would result in similar benefit, specifically providing {\em robustness} with respect to operation across a range of environmental conditions, potentially  resulting in increased longevity of the swarm. A natural question then arises: {\em how can we evolve functional diversity within a large swarm?} Existing research with Evolutionary Robotics tends towards that view that this is a challenging proposition \cite{montanier2016behavioral}, noting that very specific conditions concerning reproductive isolation must be met or at least careful choice of operators is required.However, an alternative paradigm known as {\em Quality Diversity} (QD) for evolving diversity has recently been introduced: these algorithms are designed to locate a maximally diverse collection of individuals, according to some low-dimensional set of behavioural characteristics, in which each individual is as high-performing as possible. Examples of QD algorithms include Novelty Search \cite{lehman2008exploiting} and MAP-Elites \cite{mouret2015illuminating}, with seminal applications in robotics domains, for instance, discovering a set of morphologies for  successful walking virtual creatures \cite{lehman2011evolving}, or diverse behaviours for maze-navigation \cite{pugh2016}. 


Typically, QD algorithms maintain a discrete archive (map) \\
formed by binning a low-dimensional representation of feature vectors,  with a single {\em elite}
individual stored within each bin (i.e. the best found so far with those features).  The archive can be completely external to the breeding population, having no influence over evolution \cite{pugh2016searching}. Alternatively, as in MAP-Elites \cite{mouret2015illuminating}, the archive is used to select genomes for variation and is an intrinsic part of  the evolutionary process.

Central to all the existing QD algorithms just described is the maintenance of a {\em single} archive, that is accessed and updated by a population at each generation. However, in a real-world robotic swarm, environmental conditions mean that communication is often limited to local interactions. Hence in their current form, the practical applicability of QD algorithms as a mechanism for evolving {\em swarm} diversity appears limited, given that individual members of the swarm may be unable to access or update a centralised archive.


We therefore propose a novel decentralised QD algorithm for maximising behavioural diversity with an evolving swarm where each robot swarm is tasked with the same single objective. The eventual goal of this research is to study whether this will lead to increased robustness in dynamic environments. 
The first key step, described here, is to propose and demonstrate a method that evolves the required diversity across the swarm. Our approach hybridises an open-ended distributed evolutionary algorithm mEDEA \cite{bredeche2010environment} with MAP-Elites \cite{mouret2015illuminating}  and is referred to as $EDQD$  - {\em Embodied Distributed Quality Diversity}. The algorithm simultaneously evolves multiple diverse behaviours across a swarm, through local communication only. Its goal is to evolve a swarm that maximises functional diversity, i.e. maximises the number of distinct approaches to achieving the same task. We speculate that evolving a behaviourally diverse swarm will ``future proof'' the swarm, providing robustness to a broad spectrum of potential environmental conditions --- even if some of the evolved behaviours are sub-optimal with respect to the {\em current} environment. The specific goals of the paper are as follows:


\begin{itemize}
\item  Evaluate the extent to which  a decentralised QD algorithm can be used to obtain a swarm exhibiting functional diversity w.r.t a simple foraging task
\item Evaluate the quality of each evolved behaviour in the swarm  w.r.t the estimated optimal  performance for that  behaviour  and the reliability with which high-performing behaviours are discovered


\item Compare the diversity and relative quality of behaviours evolved using the QD method to a standard distributed evolutionary algorithm (mEDEA).
\end{itemize}

The paper makes  a number of contributions.  Firstly, it describes the first completely {\em decentralised} version of a QD algorithm for {\em simultaneous evolution of multiple diverse behaviours within a swarm}. This differentiates it from existing work with robotics in which a single archive is used to evolve multiple behaviours for a {\em single} robot. Secondly, it proposes and evaluates four different methods of sharing and updating map-archives amongst the swarm. These different methods alter the amount of information shared in an attempt to maximise diversity. Experimental results show that the  method significantly outperforms a distributed EA in terms of simultaneously evolving diverse behaviours. Thirdly, it extends the usage of the QD paradigm, which is typically directed towards either {\em illumination} of a feature-space \cite{mouret2015illuminating} or to providing  a range of pre-evolved options for selecting a single  behaviour to match current environmental conditions \cite{cully2015robots}.  It represents the first step in a longer-term vision to show that a functionally-diverse swarm will be robust to environmental change, and that it can continue to learn over time --- key qualities of   practical robotic swarms.




\section{Background}

\begin{figure}
\includegraphics[width=0.47\textwidth,page=1,scale=1,trim={2cm 8.9cm 3cm 3.85cm},clip]{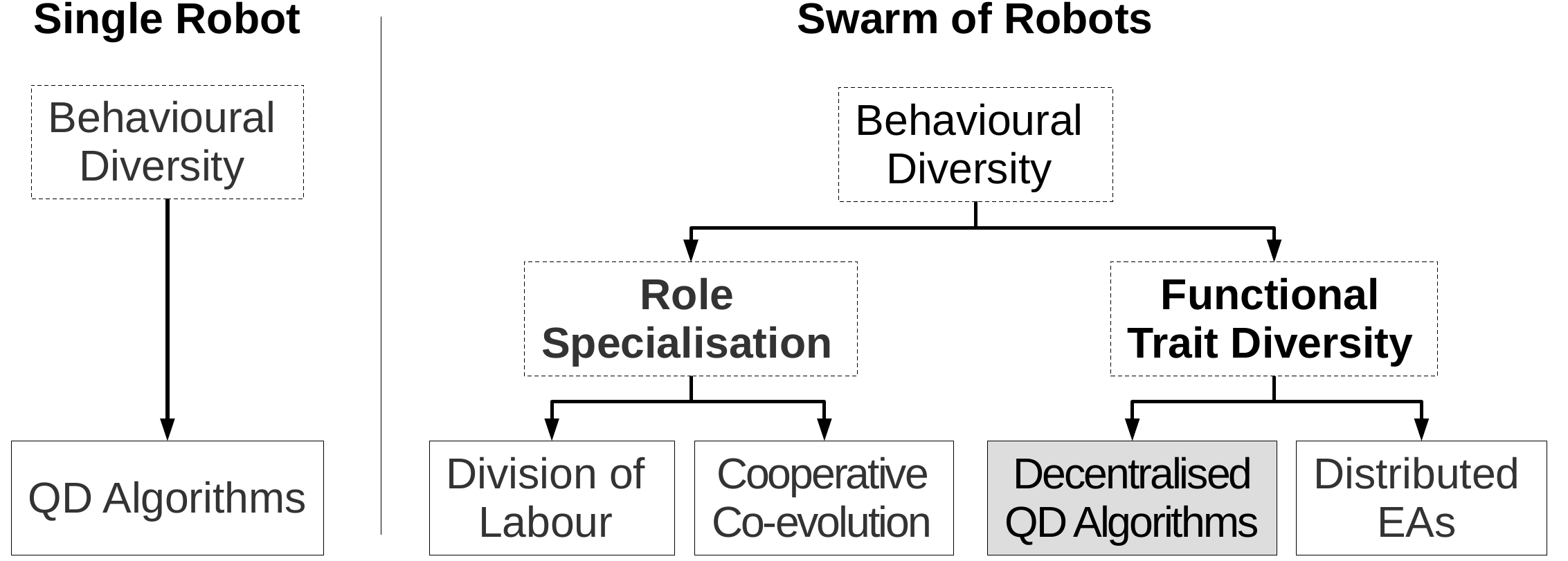}
\caption{\label{fig:tax} A taxonomy of research focusing on behavioural diversity in Evolutionary Robotics. The novel contribution of this paper is shaded in grey}
\end{figure}

As the terms {\em behavioural/functional diversity} are used in a broad range of contexts with evolutionary robotics, it is helpful to  position existing research with the field with the help of a simple taxonomy,  described in figure \ref{fig:tax}.

A clear distinction arises between work that evolves multiple distinct behaviours for a {\em single} robot and that which attempts to simultaneously evolve multiple behaviours within a {\em swarm}.  The former has been tackled extensively in recent years through the use of {\em quality-diversity} algorithms: a
 new class of algorithms that return an archive of diverse, high-quality behaviours in a single run. An overview of recent work is provided in by Pugh {\em et al} \cite{pugh2016}. On the one hand, QD algorithms  encourage exploration of the search space in order to find better solutions, overcoming deception \cite{pugh2016searching}. On the other, they can be used to develop an archive of behaviours that can be for example used in future to guide a trial-and-error learning algorithm to select  new appropriate behaviour in the face of damage \cite{cully2015robots}.

This article concerns the latter category, evolution of a {\em swarm}.  Here, a further distinction can be made between evolving {\em role specialisation} and {\em functional trait diversity} with the swarm. The first category is commonly referred to as {\em division of labour} and covers scenarios in which swarms divide into sub-groups, each accomplishing a sub-task, in order to solve a complex problem. This kind of role-specialisation, observed in social insects \cite{chittka2009learning} and more rarely in mammals \cite{gazda2005division} has been a focus of much effort with robotics and more generally in problem solving \cite{potter2000cooperative}.

However, our work is positioned within the sub-category classified as {\em functional trait diversity}: the term refers to the presence of multiple behavioural traits within a group that result in multiple strategies for achieving the same goal. Montanier {\em et al} \cite{montanier2016behavioral} find that behavioural diversity in this case is very hard to achieve, noting that reproductive isolation is necessary, whether such isolation is due to geographic constraints or particular mating strategies, and that large population sizes help. 
Haasdijk {\em et al} \cite{haasdijk2014combining} considered evolution of co-existing foraging behaviours within a population and  found they needed to introduce a market mechanism favouring sub-groups in order to evolve diversity. Trueba {\em et al.} \cite{trueba2013specialization}  conducted an in-depth empirical study of behavioural specialization within the {\em same} geographic location  and showed that judicious choice of evolutionary operators
 could be used to enforce behavioural specialisation but the study was greatly simplified in that robots simply selected from three predefined behaviours.
 
 We address the challenges just described in obtaining functional trait diversity by hybridising  MAP-Elites, a QD algorithm shown to work well in applications involving a {\em single} robot, with a distributed evolutionary algorithm for evolving behaviours within a {\em swarm}. The new algorithm --- {\em Embodied Distributed Quality Diversity} algorithm, EDQD is now described.

\section{The Embodied Distributed Quality
Diversity Algorithm (EDQD) }

\begin{table*}
\caption{\label{tab:ED-QD} Variants of the EDQD Algorithm}
\begin{tabular}{cccc}
\hline
Acronym & Creation of SelectMap & Update of MemoryMap & Update of LocalMap \\
\hline
R&random selection from ReceivedMapList & - &  Executed Genome \\
M1&merge(ReceivedMapList) & - & Executed Genome\\
M2&merge(ReceivedMapList, MemoryMap) & ReceivedMapList &  Executed Genome\\
M3&merge(ReceivedMapList, LocalMap),& - & Executed Genome, ReceivedMapList \\
\hline
\end{tabular}
\end{table*}

$EDQD$ hybridises a distributed environment-driven evolutionary adaptation algorithm  called mEDEA  \cite{bredeche2010environment} with a novel decentralised version of a  MAP-Elites algorithm \cite{mouret2015illuminating}. The mEDEA algorithm has been extensively studied in previous works and provides a good baseline for embodied evolution. In the original version of this algorithm,  there is no explicit fitness function. Therefore, no selection pressure w.r.t. fitness value is applied, but selection pressure w.r.t. ability to spread one's own genome is still at work. Later versions introduced an additional fitness mechanism \cite{Perez2014} to regulate the trade-off between the exploitation of a fitness function and the exploration of solutions allowing the survival of robots. We adopt this version here due to the fact we also introduce an explicit task.
 
In brief, in mEDEA, for a fixed period (lifetime), robots move according to their control algorithm (a neural network, specified by weights in the genome).
As they move, they broadcast their genome, which is received and stored by any robot within range.  At the end of  this period, a robot selects a random genome from its list of  collected genomes and applies a variation operator. This takes the  form of a Gaussian random mutation operator that can be easily tuned through a $\sigma$  parameter. Robots that have not collected any genomes become  inactive, thus temporarily reducing the population size.

In EDQD, each robot stores an n-dimensional discrete map defined by functional traits related to the specified task termed the {\em LocalMap} (LM).  In contrast to mEDEA in which a robot broadcasts  its current {\em genome},  in EDQD, each robot broadcasts its {\em LocalMap},
an elite archive of the genomes it has previously evaluated. All robots in range receive a copy of the broadcast map and store it in a temporary list. At the start of each  new generation, every robot selects a new genome from the maps received during the previous generation. A variation operator is applied as in mEDEA and the new controller evaluated over the defined robot-lifetime. The 
{\em LocalMap} 
is then updated based on the fitness of the evaluated 
phenome 
(where update is contingent on the discovery of a new elite). 
These fitness evaluations are inherently noisy, as in all online embodied algorithms, which arises from the fact that agents are evaluated in a shared environment and influence each other.
The algorithm is defined in Algorithm \ref{alg:me-medea}.

\begin{figure*}
\centering
	\subfloat[EDQD-R]{
		\includegraphics[page=4,scale=0.55,trim={7cm 7cm 7cm 6.2cm},clip]{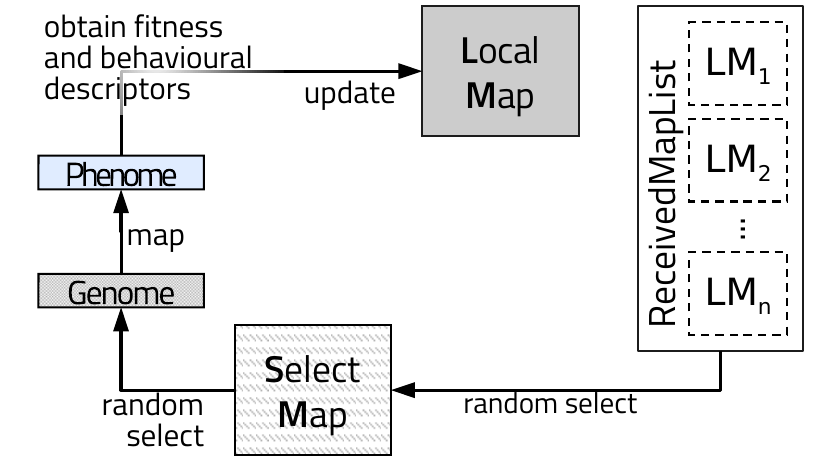}
	}
    \qquad
	\subfloat[EDQD-M1]{
		\includegraphics[page=5,scale=0.55,trim={7cm 7cm 7cm 6.2cm},clip]{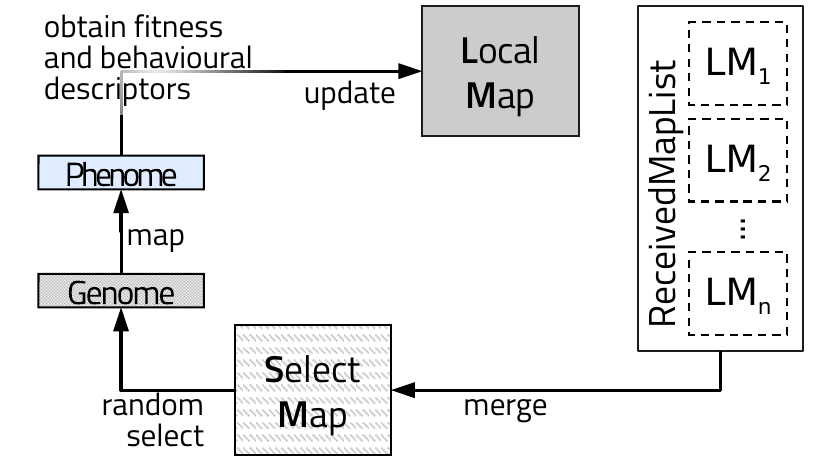}
	}
    \\
	\subfloat[EDQD-M2]{
		\includegraphics[page=6,scale=0.55,trim={7cm 7cm 7cm 6.2cm},clip]{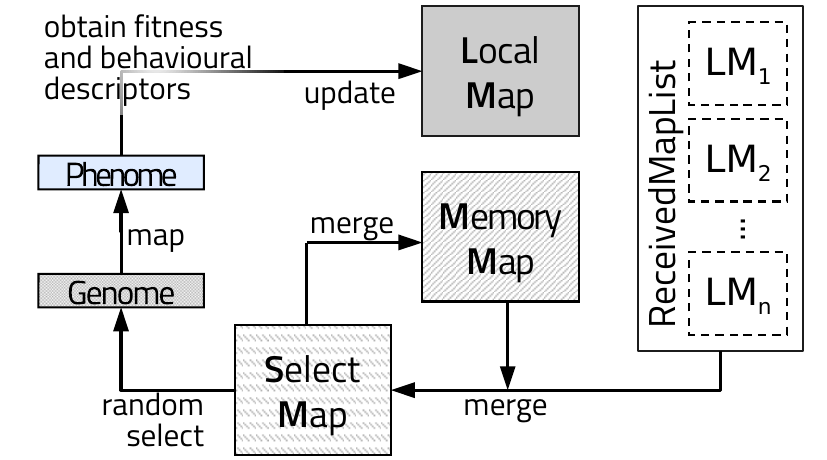}
	}
    \qquad
	\subfloat[EDQD-M3]{
		\includegraphics[page=7,scale=0.55,trim={7cm 7cm 7cm 6.2cm},clip]{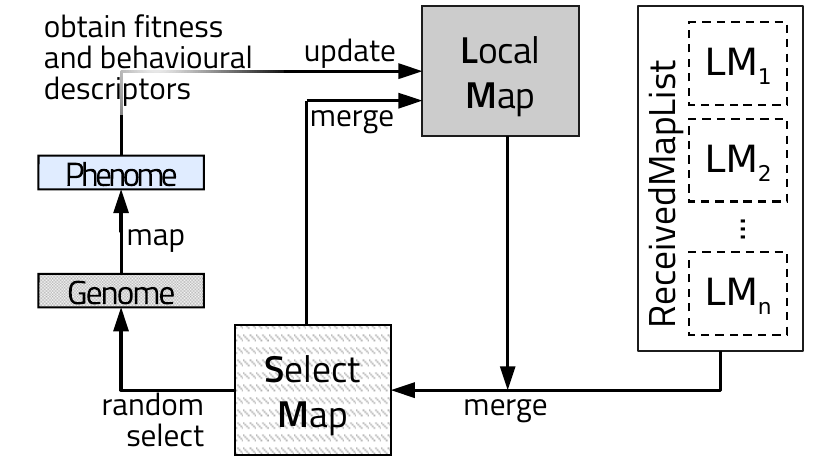}
	}
\caption{\label{fig:variants} Variants of the EDQD Algorithm}
\end{figure*}

We explore four variants of the algorithm which are differentiated by the method in which the robots select a new genome from maps they have received, and the method in which the $LocalMap$ is updated. Four variants of the EDQD algorithm are proposed. These are summarised in table \ref{tab:ED-QD} and in figure \ref{fig:variants}, and described below.  The following definitions apply throughout:

\begin{itemize}
\item {\bf Local Map}  Each robot maintains a local map ($LM$) consisting of a 2-dimensional map containing the best solutions found so far at each point in a space defined by the two dimensions of variation. Each individual $LM$ contains elites discovered by the robot by executing genomes in its own body, with the exception of a single variation (EDQD-M3, see below) in which a LM is updated with elites obtained from other robots. This map is constructed as described by Mouret {\em et al} \cite{mouret2015illuminating}, i.e following execution,  a genome with fitness $x$ is mapped to a feature vector $b$: the genome replaces the current occupant of the cell corresponding to $b$  if it has higher fitness\footnote{If the new 
phenome
has {\em equal} fitness to the current occupant, then the 
phenome
with behavioural vector $b$ closest to the cell centre is retained; this method is used in {\em code} provided by \cite{mouret2015illuminating}} (see Algorithm \ref{alg:meUpdate}). Note however, that unlike MAP-Elites, there is no initialisation phase in which a map is first created by sampling $\mathcal{G}$ random genomes.

\item {\bf Received Map List}  A set of maps collected during one lifetime through encounters

\item {\bf Select Map} At the end of each lifetime, each robot has a {\em ReceivedMapList}. We consider four methods by which the robot can condense this information into a single map: the {\em SelectMap (SM)}. A genome is selected at random from the {\em SelectMap} to be executed in the next generation.

\item {\bf Memory Map}  In variant EDQD-M2, a robot keeps an additional map in memory which combines elites from all LMs it has ever received across all generations of the algorithm.

\end{itemize}

\paragraph{{\bf EDQD-R}}
At the end of each lifetime, each robot attempts to update its own local map with the fitness of its current 
phenome,
according to Algorithm \ref{alg:meUpdate}. Following this, in similar vein to mEDEA, the robot simply selects  a random map from the {\em ReceivedMapList} which becomes the {\em SelectMap}. A random genome is then chosen from the {\em SelectMap}. The robot then empties its {\em ReceivedMapList}.

\paragraph{{\bf EDQD-M1}}
As above, each {\em LocalMap} is first updated with the result of current 
phenome. 
The robot forms the {\em SelectMap} by merging the maps contained in the {\em ReceivedMapList} (Algorithm \ref{alg:merge}).  As above, it then selects a random genome from the {\em SelectMap}, before deleting  the merged map and emptying its {\em ReceivedMapList}. This provides more selection-pressure than variant {\bf EDQD-R}, as the merged map will potentially have more cells covered than any single map and contain the genome with the highest phenotypic fitness for overlapping cells.

\paragraph{{\bf EDQD-M2}}
Similar to above, however in this case, at iteration 0 the robot creates an empty {\em MemoryMap}. At the end of each generation, a merged map is created (Algorithm \ref{alg:merge}), and this is then merged with the {\em Memory Map}. This forms the {\em SelectMap} from which the robot selects a random genome.  The {\em MemoryMap} is updated to be equivalent to the newly merged map.
Essentially, this results in the robot maintaining a list of all elites collected from every robot it meets during its lifetime that it can use to {\em select} a new genome from.   This further increases selection pressure, as the {\em MemoryMap} stores all elites known to all robots ever encountered.

\paragraph{{\bf EDQD-M3}}
In this case,  the {\em ReceivedMapList} is merged with the robot's own {\em LocalMap} to form the {\em SelectMap} and a genome selected at random from this map. The robot's {\em LocalMap} is also updated with this information, unlike in the methods above. 
This alters the information {\em broadcast} by each robot (the LM), resulting in a more global sharing of information across the population, as each robot now receives a map containing information from genomes executed by many robots. This is expected to provide most selection pressure as it combines information from elites found by an individual robots with all elites found by all encountered robots.

\begin{algorithm}
$genome$.randomInitialise()\;
$localMap$.create()\;
\While{(generations < maxGen)}{
	\For {$iteration = 0$ to $lifetime$}{
    	\If {agent.hasGenome()}{
			agent.move()\;
			broadcast($localMap$)\;
         }
    	mapList $\leftarrow receivedMaps$
	}
    
    localMap $\leftarrow$ updateLocalMap($genome, fitness$)\;
      
	$genome$.empty()\;

	\If{$mapList$.size() $> 0$}{
    	createSelectMap() \;
		$genome =$ applyVariation($select_{random}$($selectMap$))\;

    \If{collectedMapMemory.isForget()} {
        mapList.empty()\;
    }
    }
}
\caption{EDQD Algorithm: hybridised from mEDEA \cite{bredeche2010environment} and MAP-Elites \cite{mouret2015illuminating}}
\label{alg:me-medea}
\end{algorithm}

\begin{algorithm}
$x \leftarrow random\_selection(\mathcal{X})$\;
$x' \leftarrow random\_variation(x)$\;
$b' \leftarrow featureDescriptor(x')$ \;
$p' \leftarrow performance(x')$ \;

\If {$\mathcal{P}(b')$ = $\varnothing$ or $\mathcal{P}(b') < p'$ }{
$\mathcal{P}(b') \leftarrow p'$ \;
$\mathcal{X}(b') \leftarrow x'$\;
}
\Return{map($\mathcal{P},\mathcal{X}$)}
\caption{\label{alg:meUpdate} Update of cells within a {\em LocalMap} (as defined in \cite{mouret2015illuminating})}
\end{algorithm}

\begin{algorithm}
$\mathcal{M}1 \leftarrow (\mathcal{X}_1,\mathcal{P}_1)$  \;
$\mathcal{M}2 \leftarrow (\mathcal{X}_2,\mathcal{P}_2)$  \;
\ForAll{cells $c$ in map}{
\If{$\mathcal{P}_{M2}(c) > \mathcal{P}_{M1}(c)$} {
$\mathcal{P}_{M1}(c) \leftarrow \mathcal{P}_{M2}(c)$ \;
$\mathcal{X}_{M1}(c) \leftarrow \mathcal{X}_{M2}(c)$ \;
}
}
\Return{$\mathcal{M}1$}
\caption{\label{alg:merge} Merge two maps: if there is a list of maps, then process is repeated until all maps are merged}
\end{algorithm}

\section{Experiments}
We define a simple foraging task in which a population of robots is placed in a circular arena containing equal numbers of blue and red tokens.  Tokens are coloured cylinders of the same size as robots and are collected by a robot through contact. The fitness function is defined as the {\em total} number of tokens collected during a fixed interval called the {\em lifetime} (irrespective of token colour).  An archive is defined by two functional traits: (1) the maximum Euclidean distance from starting point within a lifetime and (2) the ratio of red:blue tokens collected by the robot. The first encourages diversity w.r.t the extent of the arena explored by the robot, while the second encourages diversity in the {\em type} of token foraged.  Traits can only be calculated retrospectively at the end of each lifetime. The two-dimensional map is discretised into 15 bins per dimension, therefore contains 225 discrete locations (approximately equal to the number of robots in the population).

All experiments are conducted in simulation using the Roborobo simulator (version 3) \cite{bredeche2013roborobo}. The arena has a diameter of 956 pixels. It contains 150 red  tokens and 150 blue tokens uniformly distributed throughout the environment. When a robot consumes a token, a token of the same colour is regenerated at a random location. Robots are cylindrical, with 12 sensors (7 toward the front uniformly covering $90^{\circ}$ with the remaining 5 evenly spaced around the remainder of the body) and two motors. A genome defines 126 weights of a feed-forward neural network with 63 inputs corresponding to 3 inputs for the RGB ground colour, the 5 values for each of the 12 sensors (the distance to the nearest object and whether that object is a robot, a wall or either of the 2 type of token) and 2 outputs corresponding to translational and rotational speeds. 

A fixed population size of 200 robots is used in all experiments. 
In all algorithms tested, a Gaussian mutation is used as the variation operator, in which 
$\sigma$ is initialised to $0.1$ and subsequently evolves.
Each robot evaluates its genome over a lifetime defined as 800 iterations. $EDQD$ algorithms are compared to $mEDEA_{fps}$ which uses fitness proportionate selection to select a new genome from the list of genomes collected during a lifetime; otherwise the parameters of this algorithm are identical to $EDQD$. Each treatment is allocated 1000 generations; for each treatment, 30 independent runs are performed. 

Statistical analysis was
conducted based on the method in \cite{steyven2017gecco} using a significance level of 5\%. 
The distributions of two results were checked using a Shapiro-Wilk test.
A Kruskal-Wallis rank sum test was performed to determine the p-value if one of the results followed a non-Gaussian distribution.
Otherwise Levene's test for homogeneity of variances was performed.
For unequal variances the p-value was determined using a Welch test, otherwise using an ANOVA test.

\section{Results}
The following section describes results obtained with respect to evaluating the four variants of EDQD in terms of diversity of behaviours evolved, and the relative quality of those behaviours.

\subsection{Diversity of expressed behaviours within a swarm}
\begin{figure}
\includegraphics[width=0.48\textwidth]{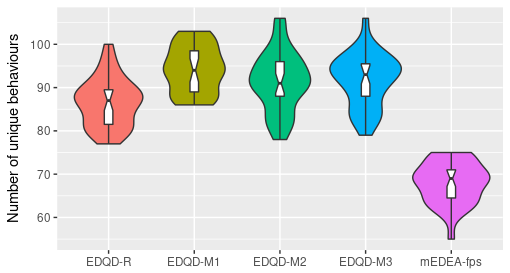}
\caption{\label{fig:swarmdiversity}Swarm Diversity: number of unique behaviours exhibited by the swarm at the end of each treatment (over 30 runs}
\end{figure}

At the end of each run (1000 generations), the behaviour of the last-executed 
phenome 
$x$ of each active\footnote{Any robot that has not collected any local-maps in the previous generation is considered as inactive} robot in the swarm is mapped to the corresponding feature descriptor $b$. A new map is constructed using algorithm \ref{alg:meUpdate}, and the number of occupied cells $n_{occ}$ counted. This is repeated for each of the 30 runs: figure \ref{fig:swarmdiversity} shows a violin plot of  the distribution of $n_{occ}$ for each treatment.
Statistical tests (table \ref{tab:uniqB}) show that  the EDQD variants all outperform $mEDEA_{fps}$; EDQD-R (that selects a random genome from a random map) is outperformed by the other three $EDQD$ variants;  however there are no significant differences between variants M1,M2,M3.

\begin{table}[ht]
\centering
\begin{tabular}{lllll}
  \hline
 & EDQD-M1 & EDQD-M2 & EDQD-M3 & mEDEA-fps \\ 
  \hline
EDQD-R & $<$ \textbf{9.21e-06} & $<$ \textbf{2.67e-03} & $<$ \textbf{1.16e-03} & $>$ \textbf{1.28e-11} \\ 
  EDQD-M1 &  & $=$ 1.05e-01 & $=$ 1.4e-01 & $>$ \textbf{1.28e-11} \\ 
  EDQD-M2 &  &  & $=$ 7.03e-01 & $>$ \textbf{1.28e-11} \\ 
  EDQD-M3 &  &  &  & $>$ \textbf{1.29e-11} \\ 
   \hline
\end{tabular}
\caption{\label{tab:uniqB} p-value and directionality obtained from pairwise comparison of number of unique behaviours exhibited by the active swarm at the final generation. Bold indicates significance at 5\% level}
\end{table}

\begin{figure}
\includegraphics[width=0.48\textwidth,trim={0.2cm 0 0 0.7cm},clip]{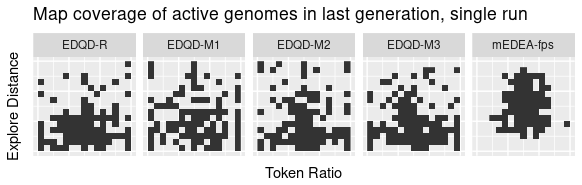}
\caption{\label{fig:swarmdActiveArchive} Map-archive of  behaviours  expressed by robots in a single swarm at the final generation (from a single run)}
\end{figure}

An alternative perspective is given in figure \ref{fig:swarmdActiveArchive} which indicates which cells in the map-archive are filled by the genomes expressed in the swarm in the final generation. 
The maps shown are generated from a single run of each treatment. Table \ref{tab:numBehaviours} shows the number of cells filled from the single run shown as well as the median cells filled per map across 30 runs. Note firstly that the EDQD variants exhibit more unique behaviours and secondly, that these behaviours are more widely {\em spread} across the behaviour space. In contrast, mEDEA$_{fps}$ tends to find a cluster of similar behaviours in the centre of the map.

\begin{table}[ht]
\centering
\begin{tabular}{rlrr}
  \hline
 & Exp. name & Count & Median Count (over 30 runs)\\ 
  \hline
  1 & EDQD-R &  86 & 87 \\ 
  2 & EDQD-M1 &  94 & 94 \\ 
  3 & EDQD-M2 &  91 & 91 \\ 
  4 & EDQD-M3 &  94 & 93 \\ 
  5 & mEDEA-fps &  71 & 69 \\ 
   \hline
\end{tabular}
\caption{\label{tab:numBehaviours} Number of unique  behaviours expressed by robots in a swarm at the final generation}
\end{table}

\begin{figure}
\includegraphics[width=0.48\textwidth,trim={0.2cm 0.1cm 0 0},clip]{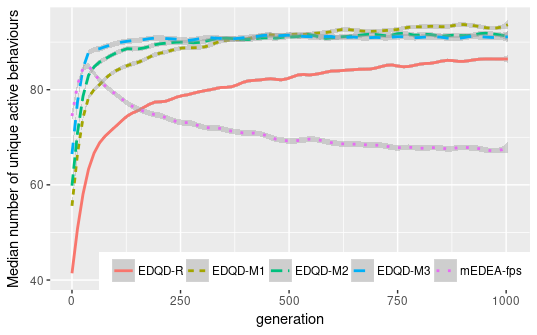}
\caption{\label{fig:swarmdiversityOneRun} Evolution of the median number  of unique behaviours exhibited by the swarm over the course of a single run
(lines have been smoothed for better visualisation)}
\end{figure}

Figure \ref{fig:swarmdiversityOneRun} shows the variation in diversity within the population (i.e. $n_{occ}$ as defined above) over the course of a single run of each treatment. The rapid loss of diversity using mEDEA is clear. In contrast, the EDQD treatments increase diversity over time. EDQD-R (select from random map) produces least diversity: the random nature of the map-selection method here limits spread of information through the population. On the other hand, the three other methods that encourage faster map propagation increase diversity rapidly.

\subsection{Quality of Evolved Behaviours}
\label{sec:quality}
As stated in section \ref{sec:intro}, the motivation of this paper is 
to determine the extent to which it is possible to evolve a diverse set of behaviours within a swarm for a single task. The previous sections clearly demonstrate that EDQD  evolves swarms that exhibit functional diversity. However, it is important to understand the impact of encouraging {\em diversity}  on the {\em quality} of behaviours discovered.

\begin{figure}
\includegraphics[width=0.48\textwidth,trim={0.2cm 0 0.3cm 0.7cm},clip]{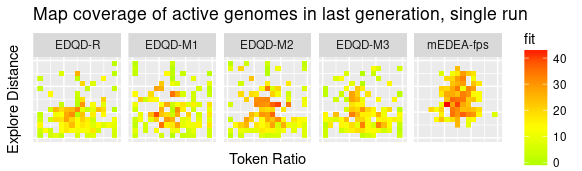}
\caption{\label{fig:swarmdActiveArchiveColour} Map-archive of  behaviours  expressed by robots in a single swarm at the final generation (from a single run), coloured according to fitness}
\end{figure}

Figure \ref{fig:swarmdActiveArchiveColour} shows the same map-archive of behaviours expressed by the swarm at the end of the final generation of a single run as shown above, but now with cells coloured according to fitness. Note that mEDEA results in a cluster of high-fitness cells in the centre of the map. The behaviours discovered by the EDQD variants, particularly around the edges of the map, are of lower fitness, although EDQD-M2 and EDQD-M3 also locate several high-performing behaviours.  For the specific environment in which experiments are conducted, it is inevitable that high-fitness behaviours are likely to occur when a controller results in a robot consuming all tokens, i.e. the  resulting token ration trait is 1:1.   A robot whose current controller results in it only reacting to red tokens and moving in (for example) tight circles
 will not collect as many tokens as a robot whose controller results in a behaviour that reacts to both types of token and moves with the same circular pattern, or indeed one that happens to explore more of its environment.
However, although such behaviours are {\em sub-optimal} in the current environment, they may be {\em optimal} in a different environment, e.g. one in which blue tokens suddenly disappear. Hence there is value in capturing them within the swarm to proof against future change.

For a given environment, it is useful to understand {\em opt-in reliability} or {\em precision} of each treatment, using the terminology introduced in \cite{mouret2015illuminating}.  For each run, if (and only if) a run creates a solution in a cell, then the precision is calculated as the average across all such cells  of the highest performing solution produced for that cell divided by the  optimal solution for that cell. 
Essentially, this reflects the
trust we can have that, if an algorithm returns a solution in a cell, that solution will be high-performing relative to {\em what is possible} for that cell. As the metric is only measured over filled-cells, it is expected that traditional objective-based evolutionary algorithms should fare well on this criterion, as they explore only a few cells but should produce high-performing solutions in those cells.

\begin{figure}
\centering
    \subfloat[EDQD variants]{
		\includegraphics[scale=0.55,trim={0 0 0 1cm},clip]{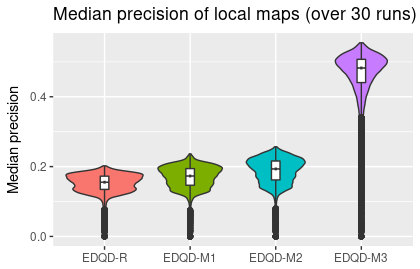}
		\vspace{-4mm}
        }
    \subfloat[mEDEA-fps]{
		\includegraphics[scale=0.55,trim={0 0 0 1cm},clip]{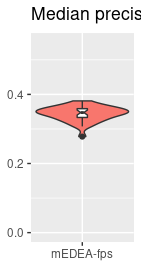}
        }
\caption{\label{fig:precisionLM} Median precision of a) each swarm at end generation (calculated as median precision of the 200 local-maps in each of 30 runs of EDQD treatments)
b) expressed genomes in last generation from mEDEA (over 30 runs)}
\end{figure}

The precision of each local map is calculated as just described.  As the optimal value is usually unknown, we follow \cite{mouret2015illuminating} in estimating this value as the best fitness found for a vector $b$ in {\em any} run of {\em any} treatment. 
The median precision per swarm of local maps is then calculated for each of the 30 runs. A violin plot of this information is given in Figure \ref{fig:precisionLM}. Although mEDEA does not produce a local map, we project the 200 fitness of the behaviours expressed by the swarm at the end generation onto a map-archive and calculate the same precision metric; this is displayed in fig. \ref{fig:precisionLM} for comparison. The results of calculating statistical significance for the EDQD treatments is also given in table \ref{tab:precisionLM}.

EDQD-M3 significantly outperforms the other EDQD variants. The median swarm precision is also higher than the precision of mEDEA across a swarm. This indicates confidence in the opt-in reliability of EDQD  in that it is capable of finding high-quality behaviours, even if some of those behaviours are sub-optimal for the current conditions.

\begin{table}[ht]
\centering
\begin{tabular}{llll}
  \hline
 & EDQD-M1 & EDQD-M2 & EDQD-M3 \\ 
  \hline
EDQD-R & $<$ \textbf{2.48e-10} & $<$ \textbf{1.34e-11} & $<$ \textbf{1.34e-11} \\ 
  EDQD-M1 &  & $<$ \textbf{1.35e-09} & $<$ \textbf{1.34e-11} \\ 
  EDQD-M2 &  &  & $<$ \textbf{1.34e-11} \\ 
   \hline
\end{tabular}
\caption{\label{tab:precisionLM} p-values obtained from pairwise comparison of median of (median) precision of 200 local-maps (over 30 runs). Bold used to indicate significance at 5\% level} 
\end{table}

\subsection{Archive of functional diversity: the swarm-map}
\begin{figure}
\includegraphics[width=0.48\textwidth,trim={0.2cm 0 0.3cm 0.7cm},clip]{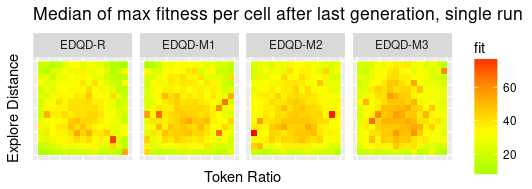}
\caption{\label{fig:swarmMap} Swarm-map: archive of best individuals from collation of all {\em LocalMaps} across the swarm of each of the EDQD algorithms at the final generation (from a single run), coloured according to fitness }
\end{figure}

The analysis conducted so far has focused on information contained within a distributed swarm as it evolves online. Given that experiments are conducted in simulation, we can extract additional information by collating the 
{\em LocalMap}s
held by the 200 individual robots at any point $t$, and merging these into a single {\em swarm-map}. It should be clear that such a map must be created externally and is {\em never} accessible to the swarm during the course of a run\footnote{Strictly speaking, it could in fact be  created by EDQD-M3 if {\em every} robot met {\em every} other robot during the course of a single lifetime}. However, in a practical scenario, it is conceivable that robots might regroup following an extended period of online evolution and that the swarm-map could be created and distributed across the swarm before evolution restarts.
The external archive would then provide  a comprehensive library of diverse behaviours for robots to draw on in future, in the face of varying environmental conditions (e.g. as described by \cite{cully2015robots} for a single robot).
Figure \ref{fig:swarmMap} shows the {\em swarm-map} created from a single run of each of the  EDQD algorithms at generation 1000. We note that {\em all} variants fill the entire archive, i.e. the swarm-map is maximally diverse. The {\em quality} of the behaviours found however appear better using variants M2 and M3.

\begin{figure}
\centering
    \subfloat[EDQD variants]{
		\includegraphics[scale=0.55,trim={0 0 0 1cm},clip]{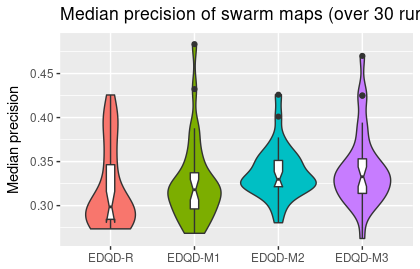}
		\vspace{-4mm}
        }
    \subfloat[mEDEA-fps]{
		\includegraphics[scale=0.55,trim={0 0 0 1cm},clip]{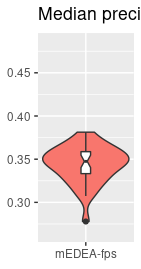}
        }
\caption{\label{fig:precisionSM} Precision of a) swarm-maps at end generation
b) active genomes in last generation (over 30 runs)}
\end{figure}

As described in section \ref{sec:quality}, the precision of a map indicates the reliability with which cells are filled with high-quality solutions. The precision of each swarm-map created from each of 30 runs is calculated and plotted in figure \ref{fig:precisionSM}(a). As in the previous section, this is also compared to the map obtained by plotting genomes expressed in robots evolved using mEDEA at the final generation (figure \ref{fig:precisionSM}(b)).
Table \ref{tab:swarmP} displays the corresponding statistical analysis obtained by comparing EDQD treatments. Combining the 200 local-maps into a single swarm-map blurs the distinction between the EDQD variants.  EDQD-R is outperformed by variants M2 and M3, and EDQD-M2 outperforms M1. Otherwise, there is no statistical difference observed.

\begin{table}[ht]
\centering
\begin{tabular}{llll}
  \hline
 & EDQD-M1 & EDQD-M2 & EDQD-M3 \\ 
  \hline
EDQD-R & $=$ 2.63e-01 & $<$ \textbf{8.29e-03} & $<$ \textbf{9.78e-03} \\ 
  EDQD-M1 &  & $<$ \textbf{2.63e-02} & $=$ 5.12e-02 \\ 
  EDQD-M2 &  &  & $=$ 9.31e-01 \\ 
   \hline
\end{tabular}
\caption{\label{tab:swarmP} p-values obtained from 
pairwise comparison of median
swarm-map precision (30 runs). Results that are significant at 5\% level shown in bold} 
\end{table}

\section{Discussion and Conclusions}
A body of evidence from both natural and computational intelligence underlines the benefits that can be derived from groups which exhibit functional diversity.  We have hypothesised that a robotic swarm might enjoy the same benefits, in terms of being {\em robust} to future changes in the environment in which it operates. If a subset of the swarm  can continue to function in the face of change (e.g. a sudden disappearance of a particular token type) then the presence  of diversity should ensure that at least a subset of the swarm can continue to function.  Furthermore, appropriate behaviours can be passed to other robots, assuming the use of an open-ended distributed evolutionary algorithm.

As a step towards achieving this goal, we have proposed a  novel decentralised quality diversity algorithm, hybridising a distributed evolutionary algorithm mEDEA with MAP-Elites --- EDQD. To the best of our knowledge, this is the first decentralised version of a QD algorithm that simultaneously evolves {\em multiple} behaviours across {\em multiple} robots. Experimental results have shown that this approach is capable of evolving a diverse swarm, with multiple behaviours being exhibited by active robots. Further to this, we have provided new insights into how to maximise diversity, showing that amalgamating maps and exploiting memory facilitates spread of information across the swarm.  We also noted that if the local archives distributed across a swarm can be aggregated into a {\em swarm-map} then a comprehensive library of potential behaviours could be obtained for future use. This approach could also be applied to evolve a library of behaviours for a single robot in {\em parallel} with the caveat that the environment each individual robot experiences is shared and manipulated by all the other robots.


The paper provides new evidence that behavioural diversity {\em can} be generated across a swarm without requiring reproduction isolation or the addition of a market-mechanism \cite{montanier2016behavioral,haasdijk2014combining}. We recognise that this comes with a caveat that functional traits  appropriate to the required task must be pre-defined, and hence the designer is required to have {\em a priori} knowledge of the traits that might be useful. However, we  believe this represents a useful first step in the quest to find a mechanism to evolve swarms that are robust to changing environments.  Recent work by Hamann \cite{Hamann} discusses how selective pressure towards diversity might be generated based on an understanding of speciation dynamics in biology without having to pre-define appropriate traits. Although this is applied to generating diverse behaviours for a single robot  navigating a maze, it may provide pointers for future work.


In contrast to some previous work in the QD domain in which archives are generated from millions of evaluations \cite{cully2015robots}, we conduct experiments using relatively few evaluations: 1000 generations and 200 robots. Mouret {\em et al} \cite{mouret2015illuminating} highlight that for a given budget of function evaluations, an EA allocates all of its evaluations to very few cells, and thus find good solutions for those cells. On the other hand,  MAP-Elites has to distribute the same budget across a much larger number cells which can result in lower fitness values being obtained. Running the EDQD treatments for more evaluations would likely improve EDQD precision, importantly without danger of convergence, due to its propensity to enforce diversity.



Finally, having shown that we can evolve a diverse swarm using EDQD, work is already under way to evaluate the central hypothesis driving this work, i.e. that the evolved swarm will be robust to dynamic changes in its environment.  A natural extension of this would be to then  show that in gradually changing environments, the repertoire of behaviours known to the swarm could gradually and continually increase, i.e. that the swarm could demonstrate lifelong learning.

\section*{Code}
Code repository: \url{https://github.com/asteyven/EDQD-GECCO2018}

\bibliographystyle{ACM-Reference-Format}
\bibliography{mapelites} 

\end{document}